%% file: ms.tex
\pdfoutput=1
\documentclass{article}

\usepackage{microtype}
\usepackage{graphicx}
\usepackage{subfigure}
\usepackage{booktabs} % for professional tables
\usepackage[table]{xcolor}
\usepackage{pgf}

\usepackage{hyperref}

\usepackage[arXiv]{icml2020}

\icmltitlerunning{Feature Synergy, Redundancy, and Independence in Global Model Explanations
using SHAP Vector Decomposition}

\usepackage{mathtools}
\usepackage{amsfonts}
\include{commands}

\begin{document}

\twocolumn[
\icmltitle{Feature Synergy, Redundancy, and Independence in Global Model Explanations
using SHAP Vector Decomposition}



\icmlsetsymbol{equal}{*}

\begin{icmlauthorlist}
\icmlauthor{Jan Ittner}{bcg}
\icmlauthor{Lukasz Bolikowski}{bcg}
\icmlauthor{Konstantin Hemker}{bcg}
\icmlauthor{Ricardo Kennedy}{bcg}
\end{icmlauthorlist}

\icmlaffiliation{bcg}{Boston Consulting Group}

\icmlcorrespondingauthor{Jan Ittner}{ittner.jan@bcg.com}

\icmlkeywords{Machine Learning, ICML}

\vskip 0.3in
]



\printAffiliationsAndNotice{}  % leave blank if no need to mention equal contribution

\begin{abstract}
We offer a new formalism for global explanations of pairwise feature dependencies and interactions in supervised models.
Building upon \shap\ values and \shap\ interaction values,
our approach decomposes feature contributions into
synergistic, redundant and independent components (S-R-I decomposition of SHAP vectors).
We propose a geometric interpretation of the components and formally prove its basic properties.
Finally, we demonstrate the utility of synergy, redundancy and independence
by applying them to a constructed data set and model.
\end{abstract}

\input{01intro}

\input{02sota}

\input{03preliminaries}

\input{04sri}

\input{05examples}

\input{06conclusions}

\appendix
\input{07appendix}

\bibliography{ms}
\bibliographystyle{icml2020}

%\appendix
%\section{Do \emph{not} have an appendix here}
%
%\textbf{\emph{Do not put content after the references.}}
%%
%Put anything that you might normally include after the references in a separate
%supplementary file.
%
%We recommend that you build supplementary material in a separate document.
%If you must create one PDF and cut it up, please be careful to use a tool that
%doesn't alter the margins, and that doesn't aggressively rewrite the PDF file.
%pdftk usually works fine. 
%
%\textbf{Please do not use Apple's preview to cut off supplementary material.} In
%previous years it has altered margins, and created headaches at the camera-ready
%stage. 
%%%%%%%%%%%%%%%%%%%%%%%%%%%%%%%%%%%%%%%%%%%%%%%%%%%%%%%%%%%%%%%%%%%%%%%%%%%%%%%
%%%%%%%%%%%%%%%%%%%%%%%%%%%%%%%%%%%%%%%%%%%%%%%%%%%%%%%%%%%%%%%%%%%%%%%%%%%%%%%

\end{document}

%% file: commands.tex
\newcommand\term[1]{\emph{#1}}

\newcommand\mvar[1]{\mathit{#1}}
\newcommand\mvec[1]{\mathbf{#1}}

\newcommand\std[1]{\|#1\|}
\newcommand\var[1]{\std{#1}^2}
\newcommand\cov[2]{\langle#1,#2\rangle}

\newcommand\feature[1]{\ensuremath{\mvar x_{#1}}}

\newcommand*\shapiv[4][]{{#2}^{#1}_{#3#4}}
\newcommand*\shapivec[4][]{\shapiv[#1]{\mvec{#2}}{#3}{#4}}
\newcommand*\shapv[3][]{\shapiv[#1]{#2}{#3}{}}
\newcommand*\shapvec[3][]{\shapivec[#1]{#2}{#3}{}}

\newcommand*\shapvalue[2][]{\shapv[#1]\phi{#2}}
\newcommand*\shapinteractionvalue[3][]{\shapiv[#1]\phi{#2}{#3}}

\newcommand*\shapvaluevec[1]{\shapvec{p}{#1}}
\newcommand*\shapinteractionvec[2]{\shapivec{p}{#1}{#2}}
\newcommand*\shapvecinter[1]{\shapvec{p'}{#1}}

\newcommand*\shapvecsynergy[2]{\shapivec{s}{#1}{#2}}
\newcommand*\shapvecautonomy[2]{\shapivec{a}{#1}{#2}}
\newcommand*\shapvecredundancy[2]{\shapivec{r}{#1}{#2}}
\newcommand*\shapvecindependence[2]{\shapivec{i}{#1}{#2}}

\newcommand*\featX{\feature i}
\newcommand*\featY{\feature j}

\newcommand*\shapX{\shapvaluevec i}
\newcommand*\shapY{\shapvaluevec j}

\newcommand*\shapXX{\shapinteractionvec i i}
\newcommand*\shapXY{\shapinteractionvec i j}

\newcommand*\shapYY{\shapinteractionvec j j}

\newcommand*\shapXYprime{\shapvecinter {ij}}

\newcommand*\synX{\shapvecsynergy i j}
\newcommand*\synY{\shapvecsynergy j i}

\newcommand*\autX{\shapvecautonomy i j}
\newcommand*\autY{\shapvecautonomy j i}

\newcommand*\redX{\shapvecredundancy i j}

\newcommand*\indX{\shapvecindependence i j}

\newenvironment{derivation}{\alignat{3}}{\endalignat}
\newcommand\derivationline[3][]{{#2} & \quad & #3 & \qquad & \text{#1}}

\newcommand\shap{\textsc{Shap}}

\newtheorem{definition}{Definition}

%% file: 01intro.tex
\section{Introduction}
\label{sec:introduction}

Understanding how and why a model produces its output is an essential part of building a robust machine learning solution.
There are various reasons why data scientists opt to ``unpack'' their models, including
\begin{enumerate}
\item Diagnostic: ensuring that good model performance is not a result of data leakage, the evaluation protocol is not compromised,
and the model has learned to properly generalise from the training data.
\item Validation: checking that relationships discovered by the model are plausible also from the perspective of domain experts
\item Feature selection: pruning redundant features with low or no marginal impact while protecting groups of synergistic features
\item Fairness and compliance: detecting a model's direct or indirect use of protected attributes to avoid discriminatory bias, or violation of other regulatory requirements
\end{enumerate}

Some machine learning models, by design, offer limited insights into their decision making process.
Examples include comparing coefficients of linear regression models,
counting how often a feature is used in random forest models,
or tracking neuron activations under various inputs in neural networks.
Still, the most valuable explanatory frameworks are those that can unpack an arbitrary ``black box'' model
without the need to access its internals.

Model explanation typically takes the form of attributing importance to input features, individually or by groups.
Several approaches have been proposed to date,
with \shap{} \cite{lundberg2017unified} being the most popular.

However, the primary focus of \shap\ is to quantify \emph{local} contributions of one or more features,
and is not designed to explain global relationships among features from the perspective of a given model:
Does the model combine information from groups of features, meaning that any feature of that group would be less impactful in the absence of its counterparts?
Which features are fully or partially redundant with respect to the target variable,
and could therefore be substituted for each other with little or no loss of model performance?

This paper offers new answers to questions such as the above,
proposing an approach with favourable mathematical properties
to quantify dependencies and interactions between features in a model:
given any pair of features $x_i$ and $x_j$, we interpret their \shap\ values across multiple observations as vectors,
then decompose them into multiple subvectors representing different types of relationships,
and quantify the strength of these relationships by the magnitudes of the vectors.
We distinguish three types of relationships: \term{synergy}, \term{redundancy}, and \term{independence}.
\begin{enumerate}
\item The \term{synergy} of feature $x_i$ relative to another feature $x_j$ quantifies the degree to which predictive contributions of $x_i$ rely on information from $x_j$.
    As an example, two features representing coordinates on a map need to be used synergistically to predict distances from arbitrary points on the map.
\item The \term{redundancy} of feature $x_i$ with feature $x_j$ quantifies the degree to which the predictive contribution of $x_i$ uses information that is also available through $x_j$.
    For example, the temperature and pressure measured in a vessel are highly redundant features since both are mutually dependent owing to the ideal gas law.
\item The \term{independence} of feature $x_i$ relative to feature $x_j$ quantifies the degree to which the predictive contribution of $x_i$ is neither synergistic or redundant with $x_j$
\end{enumerate}

Synergy, redundancy, and independence are expressed as percentages of feature importance. They are additive, and sum up to 100\% for any pair of features.
Importantly, neither relationship is necessarily symmetrical:
While one feature may replicate or complement some or all of the information provided by another feature, the reverse need not be the case.

%% file: 02sota.tex
\section{State of the Art}
\label{sec:sota}

Model interpretability is a subject of intensive research in the recent years.
However, the very notion of interpretability can be understood in different ways.
Doshi-Velez \& Kim \cite{doshi2017towards}, as well as Lipton \cite{lipton2018mythos}, and Gilpin et al. \cite{gilpin2018explaining}
worked towards clarifying related terminology, as well as listing motivations for, and flavors of, interpretability.

Pioneering works of Strumbelj \& Kononenko \cite{vstrumbelj2014explaining}
and Local Interpretable Model-agnostic Explanations (LIME) by Ribeiro et al. \cite{ribeiro2016should}
were refined into a unified framework called SHapley Additive exPlanation (\shap{}) by Lundberg \& Lee \cite{lundberg2017unified}
which is a foundation for most of the currently developed approaches.
In a follow-up article, higher-order \shap\ values, so-called \shap\ interaction values were introduced \cite{lundberg2018consistent}.
Efficient \shap\ implementations for tree ensemble models were also found \cite{lundberg2019explainable}.

As \shap\ became a reference framework for model explanation, several authors turned to exploring the utility of \shap\ and expanding it.
Rathi \cite{rathi2019generating} showed how to generate GDPR-compliant counterfactual and contrastive explanations using \shap{}.
Merrick \& Taly \cite{merrick2020explanation} demonstrated how to calculate confidence intervals of attributions.
Shapley Additive Global importancE (SAGE) \cite{covert2020understanding} were proposed
for quantifying a model's dependence on its features.
Sundararajan \& Najmi \cite{sundararajan2020many} explored axioms and desired properties of various attribution methods.

Naturally, critical analysis of \shap\ revealed its limitations.
Kumar et al. \cite{kumar2020problems} pointed to certain mathematical shortcomings of \shap\ (including the question of addressing causality)
and the fact that Shapley values represent only a summary of a game.
The same authors \cite{kumar2020shapley} offered a concept of Shapley residuals, vectors capturing information lost by Shapley values.
Their approach is based on work of Stern \& Tettenhorst \cite{stern2019hodge}, who have shown a way of decomposing an arbitrary game
and the relation of such decompositions to Shapley values.

%% file: 03preliminaries.tex
\section{Preliminaries}
\label{sec:preliminaries}

Let us start by briefly recalling the key concepts upon which the S-R-I decomposition is founded.

\subsection{Original Shapley Values}
\label{subsec:shap-values}

Shapley values were originally introduced as a concept in game theory
to describe the distribution total surplus of different coalitions of players
in an $n$-person game.
As each player in different coalitions
has a different contribution to the final outcome,
Shapley values provide a way of modeling
the marginal contribution of each player to the overall cooperation of the game.
Formally, Shapley \cite{shapley1953s} expresses the amount allocated to player $i$ in a collaborative game with players $N$ and outcomes $f_{x}(S)$ for any subset \term{(coalition)} of players $S\subseteq N$ as:
    \begin{derivation}
        \derivationline{}{
            \label{eq:shap-value}
			\shapvalue i & = \sum_{S\subseteq N\setminus \{i\}}{\frac {|S|!\;(|N|-|S|-1)!}{|N|!}}
				\nabla_i(S)
        } \\
        \derivationline{\text{where}}{
            \label{eq:model-contribution}
            \nabla_{i} & = f_x(S\cup \{i\})-f_x(S)
        }
    \end{derivation}
$\shapvalue i$~expresses the average incremental contribution of player $i$ 
when added to all possible permutations of coalitions $S\subseteq N\setminus \{i\}$.

\subsection{\shap\ Vectors}
\label{subsec:shap-vectors}

$\shap$ values are an application of Shapley values for a predictive model $f:\mathbb{R}^n\to\mathbb{R}$. 
In this context, the game outcome $f_x$ is the model evaluated for a sample $x\in\mathbb{R}^n$ with different sets of features present. 
“Players” are the features used in the model and “coalitions” of features correspond to subsets of features that are provided to the model to make predictions. 
The term $f_x(S)$ in~\eqref{eq:shap-value} is defined to be the original model $f$ restricted to use only features in $S$, by taking the expectation value over features not in $S$. 
In the notation of~\cite{chen2020true}:
    \begin{equation}
        \label{eq:cond-expectation}
        f_x(S) = \mathbb{E}[f(x)|S]
    \end{equation}
In particular,
    \begin{derivation}
        \derivationline{}{
            f_x(N) & = \mathbb{E}[f(x)|N] = f(x)
        } \\
        \derivationline{\text{and}}{
            f_x(\emptyset) & = \mathbb{E}[f(x)|\emptyset] = \mathbb{E}[f(x)]
        }
    \end{derivation}
Given $M$ samples in the training corpus for the model,
we can calculate the \shap\ value for each feature of each sample,
resulting in a $N \times M$ \shap\ value matrix
for each feature~$\featX$ and observation~$u$.

In turn, we define the \term{\shap\ vector} as
    \begin{equation}
        \label{eq:shapvector}
        \shapvaluevec i = (\shapvalue[1]i, \dots, \shapvalue[m]i)
    \end{equation}
being the \shap\ values for samples~$u = 1\dots m$ for feature~$i$.

\subsection{$\shap$ Interaction Vectors}
\label{subsec:shap-int-vectors}

$\shap$ interaction effects \cite{lundberg2018consistent}
quantify the interactions between any pair of features $x_i$ and $x_j$
by calculating the difference between
the $\shap$ value for feature~$i$ when $j$ is present,
and the $\shap$ value for feature~$i$ when $j$ is absent.
Formally, this relationship is captured by $\nabla_{ij}$
in~\eqref{eq:interaction_value} and~\eqref{eq:interaction_value_2}.
\begin{align}
    \label{eq:interaction_value}
    \shapinteractionvalue i j &= \sum_{
        S \subseteq N \setminus \{i, j\}
    }{
        \frac{
            |S|! (|N| - |S| - 2)!
        }{
            2(|N|-1)!
        }
        \nabla_{ij}(S)
    } \\
    \label{eq:interaction_value_2}
    \nabla_{ij} &= f_x(S \cup \{i,j\}) - f_x(S \cup \{i\}) \nonumber \\
    & - (f_x(S \cup \{j\}) - f_x(S))
\end{align}
where $S$ is a coalition of features representing a subset of all features $N$.
The summation extends over all possible coalitions of $N$
that don't contain the feature pair~$\{i,j\}$.
In~\eqref{eq:interaction_value} the $\shap$ interaction value
is split equally between features $x_i$ and $x_j$
hence $\shapinteractionvalue i j = \shapinteractionvalue j i$.
We can isolate the \term{main effect} $\shapinteractionvalue i i$ for feature $x_i$ by subtracting the interaction values
for all $j \neq i$ from the \shap\ value $\shapvalue i$:
\begin{align}
    \label{eq:main_effect}
    \shapinteractionvalue i i &= \shapvalue i - \sum_{j \neq i} \shapinteractionvalue i j
\end{align}

Similarly to \term{$\shap$ vectors},
we define the \term{\shap\ interaction vector}
as the vector of \shap\ values for samples~$u \in 1...m$
given a pair of features ~$\{x_i, x_j\}$:
\begin{derivation}
    \label{eq:shap_interaction_vector}
    \derivationline{}{
        \shapXY & = (\shapinteractionvalue[1]i j, \dots, \shapinteractionvalue[m]i j ) \quad \forall i, j \in N \times N
    } 
\end{derivation}

From~\eqref{eq:main_effect} in conjunction with~\eqref{eq:shapvector} 
and~\eqref{eq:shap_interaction_vector}, it follows that 
all interaction vectors for feature $x_i$
add up to the \shap\ vector for $x_i$:
    \begin{derivation}
        \derivationline{}{
            \shapX & = \sum_{j\in N}\shapXY \quad \forall i
        }
    \end{derivation}

%% file: 04sri.tex
\section{Synergy, Redundancy, and Independence}
\label{sec:sri}

In the following section we will introduce and examine various $m$-dimensional vectors,
where $m$ is the number of observations.
Vectors representing \shap\ values \eqref{eq:shapvector} and \shap\ interaction values \eqref{eq:shap_interaction_vector}
will be our building material,
from which we will construct other informative vectors.
Without loss of generality, at all times,
we will focus on one feature, $\featX$ (with corresponding \shap\ vector $\shapX$),
and explore its relationship with one other feature $\featY$ 
(with \shap\ vector $\shapY$ and \shap\ interaction vector $\shapXY$).

\begin{figure}[ht]
\vskip 0.2in
\begin{center}
\centerline{\includegraphics[width=\columnwidth]{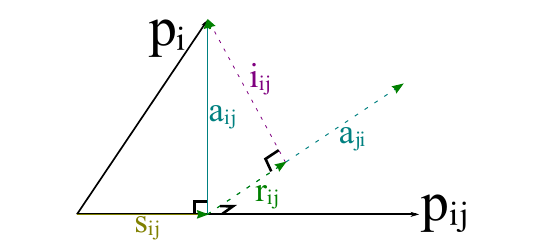}}
\caption{%
Geometric interpretation
of synergy, redundancy and independence
of feature $\featX$ relative to feature $\featY$.
In this 3-dimensional representation,
vectors $\shapX$, $\shapXY$, $\synX$ and $\autX$ are co-planar
and in the plane of the paper.
Vectors $\autX$, $\autY$, $\redX$ and $\indX$ are co-planar
and in a plane orthogonal to the paper (for better visibility, the 
perspective is slightly skewed sideways).
Feature vector $\shapX$ is projected on interaction vector $\shapXY$
to obtain synergy vector $\synX$.
Autonomy vector $\autX$ is orthogonal to $\synX$
and the two add up to $\shapX$.
Redundancy vector $\redX$ is a projection of $\autX$ onto $\autY$
($\autY$ is the autonomy vector
from the perspective of feature $\featY$).
Independence vector $\indX$ is orthogonal to $\redX$
and the two add up to~$\autX$.
}
\label{fig:geometric-interpretation}
\end{center}
\vskip -0.2in
\end{figure}

We will be concerned with angles between vectors in the $m$-dimensional space.
The smaller the angle between two vectors, the more information is shared by them.
Our goal will often be to decompose vectors into orthogonal components
(see Figure \ref{fig:geometric-interpretation}).

\subsection{Vector Representation}

\begin{definition}[Synergy vector]
	\label{def:synergy}
	\begin{equation}
	     \label{eq:synergy}
     	\synX = \frac{\cov{\shapX}{\shapXY}}{\var{\shapXY}}
		\shapXY
		\quad \forall i \neq j
	\end{equation}
\end{definition}

Geometrically speaking, the synergy vector for $\featX$ and $\featY$
is a projection of $\shapX$ on $\shapXY$.
Synergy represents the advantage that feature $\featX$ receives when aided by $\featY$.

For example, if features $\featX$ and $\featY$ represent
geographic latitude and longitude,
and our function is elevation above mean sea level,
then both features work synergistically
and neither can determine the outcome without the other.

Note that the definition is asymmetric, hence $\synY$ need not equal $\synX$.

\begin{definition}[Autonomy vector]
    \label{def:autonomy}
    \begin{equation}
        \label{eq:autonomy}
        \autX = \shapX - \synX \quad \forall i \neq j
	\end{equation}
\end{definition}

Autonomy is the converse of synergy.
As such, autonomy represents the predictive contributions $\featX$ makes without help from $\featY$,
either because it is redundant, or independent (subsequent definitions will help us distinguish between these two cases).

Geometrically, the autonomy vector is perpendicular to the synergy vector,
and both add up to $\shapX$.

\begin{definition}[Redundancy vector]
    \label{def:redundancy}
    \begin{equation}
        \label{eq:redundancy}
        \redX = \frac{\cov{\autX}{\autY}}{\var{\autY}} \autY
        \quad \forall i \neq j
	\end{equation}
\end{definition}
The redundancy vector represents information in $\featX$ that is replicated by $\featY$.
Geometrically, this is the projection of vector $\autX$ onto vector $\autY$.

For example, distance in kilometres and distance in miles are perfectly redundant features,
whereas a child's age and height are partially (but not fully) redundant.

\begin{definition}[Independence vector]
    \label{def:independence}
    \begin{equation}
        \label{eq:independence}
        \indX = \autX - \redX
        \quad \forall i \neq j
    \end{equation}
\end{definition}
Independence represents the information in feature $\featX$
that has no synergy or redundancy with feature $\featY$.
Geometrically, $\indX$ and $\redX$ are orthogonal, and together they add up to $\autX$.

Let us sum up basic properties of the vectors introduced above.
First of all, it follows directly from the definitions that:
    \begin{align}
        \shapX & = \synX + \autX = \synX + \redX + \indX \\
        & \synX \perp \redX \perp \indX \perp \synX
    \end{align}

Thanks to the above, we also have:
    \begin{align}
    	\var{\shapX} & = \var{\synX} + \var{\autX}
    	= \var{\synX} + \var{\redX} + \var{\indX}
    \end{align}
For any $\featX$ and $\featY$,
the vectors
$\shapX$, $\shapXY$, $\synX$ and $\autX$
are co-planar.
Another important plane, orthogonal to the first one,
contains the vectors
$\autX$, $\autY$, $\redX$ and $\indX$
(also see Figure \ref{fig:geometric-interpretation}).

\subsection{Scalar Representation and $S$, $R$, $I$ Values}
\label{subsec:scalars}

For practical reasons, instead of working with the full vectors,
we introduce their scalar counterparts.
For each of the scalar values $S_{ij}$, $R_{ij}$ and $I_{ij}$
we have three equivalent characterisations:
\begin{itemize}
\item geometrically, as the relative length of the projection onto $\shapX$,
\item as the ratio of squared norms $\frac{\var\cdot}{\var\shapX}$,
\item as the square of the uncentered correlation coefficient $\frac{\cov{v}{w}^2}{\var{v}\var{w}}$.
\end{itemize}

\begin{definition}[Synergy value]
    \label{def:synergy}
    \begin{equation}
        \label{eq:synergy}
        S_{ij} = \frac{\cov{\synX}{\shapX}}{\var{\shapX}}
        = \frac{\var{\synX}}{\var{\shapX}}
        = \frac{\cov{\shapX}{\shapXY}^2}{\var{\shapX}\var{\shapXY}}
        \quad \forall i \neq j
    \end{equation}
\end{definition}

\begin{definition}[Redundancy value]
    \label{def:redundancy}
    \begin{equation}
        \label{eq:redundancy}
        R_{ij} = \frac{\cov{\redX}{\shapX}}{\var{\shapX}}
        = \frac{\var{\redX}}{\var{\shapX}}
        = (1 - S_{ij}) \frac{\cov{\autX}{\autY}^2}{\var{\autX}\var{\autY}}
        \quad \forall i \neq j
    \end{equation}
\end{definition}

\begin{definition}[Independence value]
    \label{def:independence}
    \begin{equation}
        \label{eq:independence}
        I_{ij} = \frac{\cov{\indX}{\shapX}}{\var{\shapX}}
        = \frac{\var{\indX}}{\var{\shapX}}
        = 1 - S_{ij} - R_{ij}
        \quad \forall i \neq j
    \end{equation}
\end{definition}

In the appendix, we derive the equivalence between the three characterizations for each scalar value in eqs.~\eqref{eq:synergy}, \eqref{eq:redundancy}, and~\eqref{eq:independence} respectively.

We have thus defined scalar values quantifying
synergy, redundancy and independence from a global perspective.
The three values are non-negative and sum up to unity:
    \begin{align}
        S_{ij} + R_{ij} + I_{ij} = 1 \\
        0 \leq S_{ij} \leq 1 \\
        0 \leq R_{ij} \leq 1 \\
        0 \leq I_{ij} \leq 1
    \end{align}

\subsection{Orthogonality Correction}
\label{subsec:orthogonality}

\shap\ interaction vectors representing
main effects $\shapXX$ 
are not guaranteed to be orthogonal to
pairwise interaction vectors $\shapXY$.
In order to split the main effects from the interaction vectors,
we correct \shap\ interaction values
by projecting them onto the subspace that is orthogonal
to $\shapXX$ and $\shapYY$.
In other words, we determine constants $\alpha$ and $\beta$ such that
\begin{align}
& \shapXYprime := \shapXY - \alpha \shapXX - \beta \shapYY \\
& \shapXX \perp \shapXYprime \perp \shapYY
\end{align}
and apply the S-I-R calculations based on the corrected vectors $\shapXYprime$.
A further formalisation of this preprocessing step is part of our current  research (see also the outlook in section~\ref{sec:conclusions}).

%% file: 05examples.tex
\section{Experimental Results}
\label{sec:experiment}

\newcommand{\gr}[1]{%
  \pgfmathsetmacro{\PercentColor}{100.0*(#1)}
  \xdef\PercentColor{\PercentColor}%
  \cellcolor{gray!\PercentColor}{#1}
}

\begin{table}[t]
\caption{Synergy, redundancy and independence values
for pairs of features of the examined model.}
\label{tab:sri}
\vskip 0.15in
\begin{center}
\begin{small}
\begin{sc}
\begin{tabular}{lccccc}
\toprule
$S_{ij}$ & $x_1$ & $x_2$ & $x_3$ & $x_4$ & $x_5$ \\
\midrule
$x_1$    &  -   & \gr{1.00} & \gr{1.00} & \gr{0.00} & \gr{0.00} \\
$x_2$    & \gr{0.79} &  -   & \gr{0.00} & \gr{0.00} & \gr{0.00} \\
$x_3$    & \gr{0.79} & \gr{0.00} &  -   & \gr{0.00} & \gr{0.00} \\
$x_4$    & \gr{0.00} & \gr{0.00} & \gr{0.00} &  -   & \gr{0.00} \\
$x_5$    & \gr{0.00} & \gr{0.00} & \gr{0.00} & \gr{0.00} &  -   \\
\bottomrule
\toprule
$R_{ij}$ & $x_1$ & $x_2$ & $x_3$ & $x_4$ & $x_5$ \\
\midrule
$x_1$    &  -   & \gr{0.00} & \gr{0.00} & \gr{0.00} & \gr{0.00} \\
$x_2$    & \gr{0.00} &  -   & \gr{1.00} & \gr{0.00} & \gr{0.00} \\
$x_3$    & \gr{0.00} & \gr{1.00} &  -   & \gr{0.00} & \gr{0.00} \\
$x_4$    & \gr{0.00} & \gr{0.00} & \gr{0.00} &  -   & \gr{0.00} \\
$x_5$    & \gr{0.00} & \gr{0.00} & \gr{0.00} & \gr{0.00} &  -   \\
\bottomrule
\toprule
$I_{ij}$ & $x_1$ & $x_2$ & $x_3$ & $x_4$ & $x_5$ \\
\midrule
$x_1$    &  -   & \gr{0.00} & \gr{0.00} & \gr{1.00} & \gr{1.00} \\
$x_2$    & \gr{0.21} &  -   & \gr{0.00} & \gr{1.00} & \gr{1.00} \\
$x_3$    & \gr{0.21} & \gr{0.00} &  -   & \gr{1.00} & \gr{1.00} \\
$x_4$    & \gr{1.00} & \gr{1.00} & \gr{1.00} &  -   & \gr{1.00} \\
$x_5$    & \gr{1.00} & \gr{1.00} & \gr{1.00} & \gr{1.00} &  -   \\
\bottomrule
\end{tabular}
\end{sc}
\end{small}
\end{center}
\vskip -0.1in
\end{table}

Let us now examine how S-R-I decomposition works in practice to gain a deeper understanding of the relationships between model features.

\begin{figure}[htb]
\vskip 0.2in
\begin{center}
\centerline{\includegraphics[width=\columnwidth]{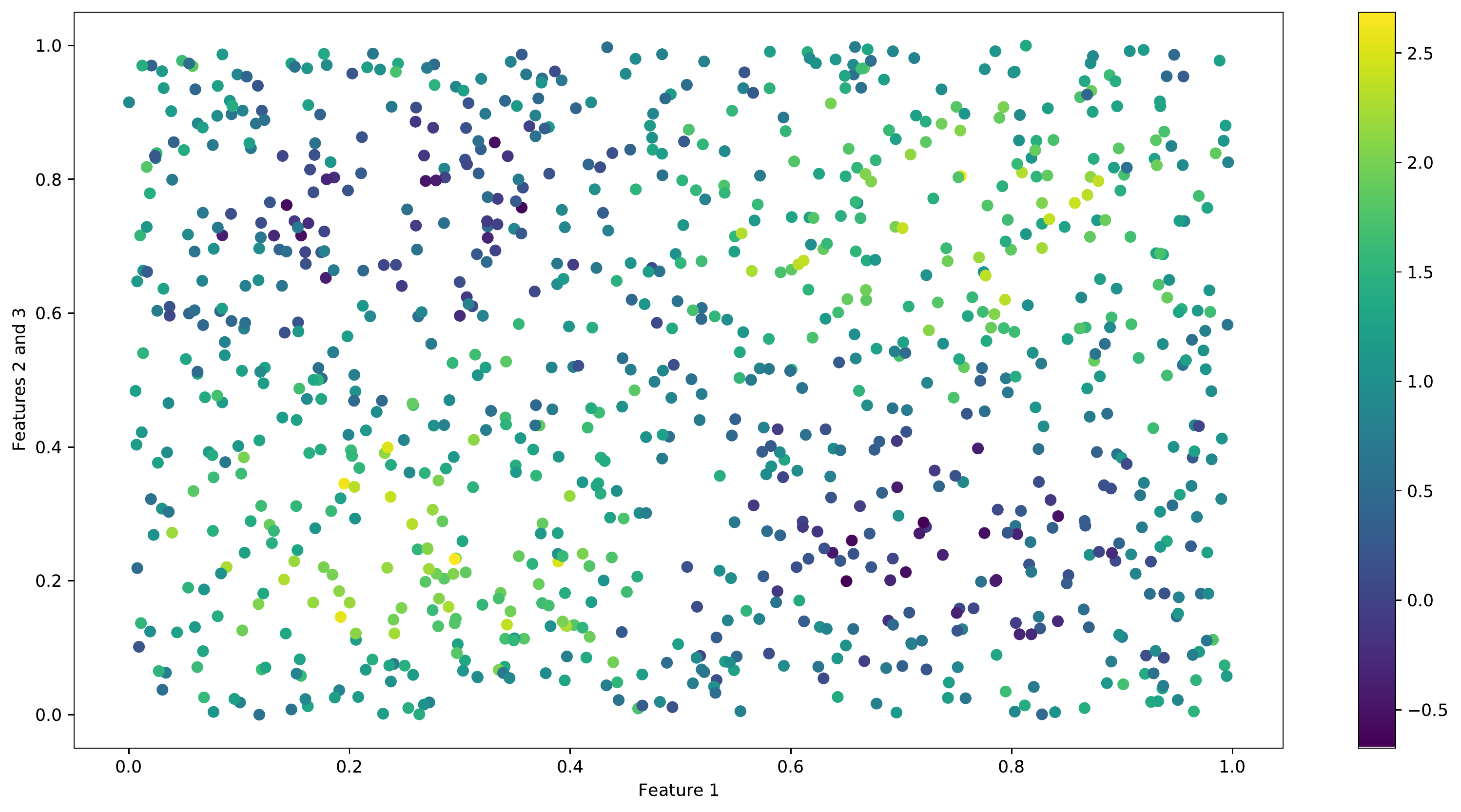}}
\caption{%
Function used in the experiment, plotted against the first feature on the x-axis,
and the second and third features (duplicated) on the y-axis.
}
\label{fig:experiment}
\end{center}
\vskip -0.2in
\end{figure}

Consider $m = 1\,000$ observations of $n = 5$ features,
represented by $m$-dimensional vectors: $\{\mvec{x}_1, \dots, \mvec{x}_n\}$.
Each value of $\mvec{x}_1$, $\mvec{x}_2$, $\mvec{x}_4$ and $\mvec{x}_5$
is drawn independently from uniform distribution $[0, 1]$,
while $\mvec{x}_3 = \mvec{x}_2$.
Consider a model (see Figure \ref{fig:experiment}):
\begin{align}
f(\mvec{x}) := & \sin (2\pi \mvec{x}_1) \sin (2\pi \frac{\mvec{x}_2 + \mvec{x}_3}{2})
	+ \mvec{x}_4 + \mvec{x}_5
\end{align}

In other words, features $\mvec{x}_2$ and $\mvec{x}_3$
are identical, {\em redundant} copies.
Features $\mvec{x}_4$ and $\mvec{x}_5$ impact the model
{\em independently} of each other and of any other feature.
Impact of feature $\mvec{x}_1$ is linked to that of features $\mvec{x}_2$, $\mvec{x}_3$,
as neither can increase the function's value without ``co-operation'' with the others
(there is a large degree of {\em synergy} between them).

We have calculated exact \shap{} values for each observation,
applied orthogonality correction described in \ref{subsec:orthogonality},
and then calculated S-R-I decomposition for feature pairs.
Table \ref{tab:sri}
presents synergy, redundancy and independence values for each pair of features.

Investigating the results we notice that $S_{12} = S_{13} = 1$,
indicating that $\mvec{x}_1$ can provide the ``missing piece of information''
to $\mvec{x}_2$ and $\mvec{x}_3$.
At the same time, $S_{21} = S_{31} = 0.79$,
meaning that $\mvec{x}_2$ can also reinforce $\mvec{x}_1$,
but is limited by $\mvec{x}_3$
(and vice versa).

Looking at $R_{ij}$, the only pair of redundant features is $\mvec{x}_2$ and $\mvec{x}_3$, with $R_{23} = R_{32} = 1$.
We have $I_{4 \cdot} = I_{\cdot 4} = I_{5 \cdot} = I_{\cdot 5} = 1$,
expressing the fact that the last two features contribute fully independently to the overall outcome.
Lastly, as expected, in all cases $S_{ij} + R_{ij} + I_{ij} = 1$.

To sum up, we have observed that synergy, redundancy and independence values,
as defined in this paper,
are intuitive and quantifiable reflections of their respective notions.

%% file: 06conclusions.tex
\section{Conclusions}
\label{sec:conclusions}

In this work we have shown that an interaction between
any two features in a model
can be decomposed into three components:
synergy (S), redundancy (R) and independence (I).
We have characterized S-R-I using geometric properties,
and have proven equivalence between alternative formulations.
We have also used an example using a synthetic dataset to demonstrate
how a global explanation using S-R-I decomposition can enhance our understanding of the relationships among model features.

The three values are defined in terms of \shap\ values and \shap\ interaction values.
They can be efficiently calculated, so that the marginal
cost of the S-R-I decomposition is negligible.
We have released an open-source implementation of S-R-I decomposition
in our Explainable AI software library FACET: \url{https://github.com/BCG-Gamma/facet}.

The notion of global explanations using orthogonal vectors
in the space of observations deserves further attention.
Our current research focuses on determining desirable
geometric properties of interaction values, and proposing
relevant orthogonalisation steps.

%% file: 07appendix.tex
\section*{Appendix}
\label{sec:appendix}

As discussed in section~\ref{subsec:scalars}, each scalar value for synergy, 
redundancy and independence has three equivalent characterizations:
\begin{itemize}
\item geometrically, as the relative length of the projection onto $\shapX$,
\item as the ratio of squared norms $\frac{\var\cdot}{\var\shapX}$,
\item as the square of the uncentered correlation coefficient $\frac{\cov{v}{w}^2}{\var{v}\var{w}}$.
\end{itemize}

Here, we derive the equivalence between the three characterizations for each scalar value as stated for
$S_{ij}$ in eq.~\eqref{eq:synergy}, for $R_{ij}$ in eq.~\eqref{eq:redundancy}, and for $I_{ij}$ in eq.~\eqref{eq:independence} respectively.

Starting with $S_{ij}$, the equivalence in eq.~\eqref{eq:synergy} can be shown as follows:
\begin{derivation}
    \derivationline{}{
        \frac{\cov{\synX}{\shapX}}{\var{\shapX}}
        & = \frac{\cov{\synX}{\synX + \autX}}{\var{\shapX}}
        = \frac{\cov{\synX}{\synX}}{\var{\shapX}}
        = \frac{\var{\synX}}{\var{\shapX}}
    }
\end{derivation}
\begin{derivation}
    \derivationline{}{
        \frac{\cov{\synX}{\shapX}}{\var{\shapX}}
        & = \frac{\cov{\shapXY}{\shapX}}{\var{\shapX}}
            \frac{\cov{\shapX}{\shapXY}}{\var{\shapXY}}
    = \frac{\cov{\shapX}{\shapXY}^2}{\var{\shapX}\var{\shapXY}}
    }
\end{derivation}

For $R_{ij}$, the equivalence in eq.~\eqref{eq:redundancy} is due to:
\begin{derivation}
    \derivationline{}{
        \frac{\cov{\redX}{\shapX}}{\var{\shapX}}
        & = \frac{\cov{\redX}{\synX + \redX + \indX}}{\var{\shapX}} \nonumber
    }\\
    \derivationline{}{
        & = \frac{\cov{\redX}{\redX}}{\var{\shapX}}
        = \frac{\var{\redX}}{\var{\shapX}}
    }
\end{derivation}
\begin{derivation}
    \derivationline{}{
        \frac{\cov{\redX}{\shapX}}{\var{\shapX}}
        & = \frac{\cov{\shapX}{\autY}}{\var{\shapX}}
            \frac{\cov{\autX}{\autY}}{\var{\autY}} \nonumber
    }\\
    \derivationline{}{
        & = \frac{\cov{\synX + \autX}{\autY}}{\var{\shapX}}
            \frac{\cov{\autX}{\autY}}{\var{\autY}} \nonumber
    }\\
    \derivationline{}{
        & = \frac{\cov{\autX}{\autY}}{\var{\shapX}}
            \frac{\cov{\autX}{\autY}}{\var{\autY}} \nonumber
    }\\
    \derivationline{}{
        & = \frac{\var{\autX}}{\var{\shapX}}
            \frac{\cov{\autX}{\autY}^2}{\var{\autX}\var{\autY}} \nonumber
    }\\
    \derivationline{}{
    & = (1 - \frac{\var{\synX}}{\var{\shapX}})
        \frac{\cov{\autX}{\autY}^2}{\var{\autX}\var{\autY}} \nonumber
    }\\
    \derivationline{}{
	& = (1 - S_{ij}) \frac{\cov{\autX}{\autY}^2}{\var{\autX}\var{\autY}}
    }
\end{derivation}

For $I_{ij}$, the equivalence in eq.~\eqref{eq:independence} is due to:
\begin{derivation}
    \derivationline{}{
        \frac{\cov{\indX}{\shapX}}{\var{\shapX}}
        & = \frac{\cov{\indX}{\synX + \redX + \indX}}{\var{\shapX}} \nonumber
    }\\
    \derivationline{}{
        & = \frac{\cov{\indX}{\indX}}{\var{\shapX}}
        = \frac{\var{\indX}}{\var{\shapX}}
    }
\end{derivation}
\begin{derivation}
    \derivationline{}{
        \frac{\var{\indX}}{\var{\shapX}}
        & = \frac{\var{\shapX} - \var{\synX}
        - \var{\redX}}{\var{\shapX}}
	 = 1 - S_{ij} - R_{ij}
    }
\end{derivation}